\documentclass[final]{cvpr}

\usepackage{times}
\usepackage{epsfig}
\usepackage{graphicx}
\usepackage{amsmath}
\usepackage{amssymb}
\usepackage{capt-of}
\usepackage{multirow}

\usepackage{color}

\newcommand{\keypoint}[1]{\vspace{0.1cm}\noindent\textbf{#1}\quad}
\newcommand{\cut}[1]{}
\DeclareMathAlphabet\mathbfcal{OMS}{cmsy}{b}{n}

\usepackage[pagebackref=true,breaklinks=true,colorlinks,bookmarks=false]{hyperref}

\def\cvprPaperID{3951} 
\def\confYear{CVPR 2021}

\begin{document}

\title{PQA: Perceptual Question Answering}

\author{Yonggang Qi$^{1}$\thanks{Equal contribution} \quad Kai Zhang$^{1*}$ \quad Aneeshan Sain$^{2}$ \quad Yi-Zhe Song$^{2}$\\
$^{1}$Beijing University of Posts and Telecommunications, CN \quad $^{2}$SketchX, CVSSP, University of Surrey, UK \\
{\tt\small \{qiyg, forer\}@bupt.edu.cn}  \qquad {\tt\small  \{a.sain, y.song\}@surrey.ac.uk}
}

\maketitle

\begin{abstract}
Perceptual organization remains one of the very few established theories on the human visual system. It underpinned many pre-deep seminal works on segmentation and detection, yet research has seen a rapid decline since the preferential shift to learning deep models. Of the limited attempts, most aimed at interpreting complex visual scenes using perceptual organizational rules. This has however been proven to be sub-optimal, since models were unable to effectively capture the visual complexity in real-world imagery. In this paper, we rejuvenate the study of perceptual organization, by advocating two positional changes: (i) we examine purposefully generated synthetic data, instead of complex real imagery, and (ii) we ask machines to synthesize novel perceptually-valid patterns, instead of explaining existing data. Our overall answer lies with the introduction of a novel visual challenge -- the challenge of perceptual question answering (PQA). Upon observing example perceptual question-answer pairs, the goal for PQA is to solve similar questions by generating answers entirely from scratch (see Figure \ref{fig: figure1}). Our first contribution is therefore the first dataset of perceptual question-answer pairs, each generated specifically for a particular Gestalt principle. We then borrow insights from human psychology to design an agent that casts perceptual organization as a self-attention problem, where a proposed grid-to-grid mapping network directly generates answer patterns from scratch. Experiments show our agent to outperform a selection of naive and strong baselines. A human study however indicates that ours uses astronomically more data to learn when compared to an average human, necessitating future research (with or without our dataset). 
\end{abstract}

\section{Introduction}

\begin{figure}
    \label {fig: figure1}
    \centering
    \includegraphics[width=\linewidth]{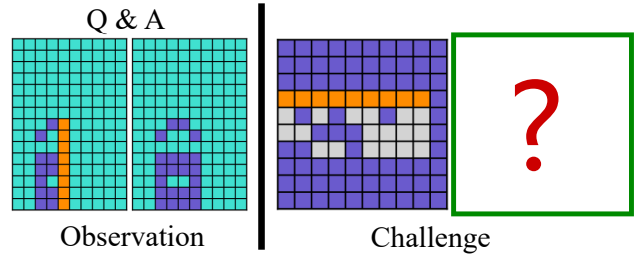}
    \caption{Perceptual Question and Answer (PQA). Given an exemplar PQA pair (Left), a new question (right) is required to be addressed, i.e.~generate answer-grid from scratch. }
    \label{fig:front}
\end{figure}

The ultimate goal for computer vision, as one might argue, lies with imitating the human visual system. Machine vision systems in their infancy were largely inspired by theories on visual perception. Perceptual organization in particular played an important part, leading to the state-of-the-art performances on image segmentation \cite{gupta2013perceptual,kootstra2011fast}, contour detection \cite{arbelaez2010contour,papari2008adaptive} or shape parsing \cite{macrini2011bone,li2017deep} for example. Research efforts have diminished thereafter with the introduction of deep learning, where the need for modeling human visual systems are no longer deemed necessary, as performances on certain vision tasks have already reached/exceeded human level \cite{krizhevsky2017imagenet,mnih2015human,he2016deep}.

Perceptual organization theory essentially boils down to a finite set of rules (called Gestalt) that collectively guide the reasoning of our visual system \cite{wertheimer1938gestalt,wagemans2015historical,wagemans2012century}. Cognitive scientists have been mostly interested in computationally modeling these rules \cite{zhu1999embedding,elder2002ecological}, and reason using them in a collective manner \cite{wang2019deepflux,greff2016tagger}. These models are however very limited in their power to represent the complex underlying mechanism of perceptual organization. Computer vision research took the opposite route, where the focus had been on using the very limited knowledge of perceptual organization to explain complex imagery \cite{qi2015making}. Yet, performances reported are largely sub-optimal and only applicable to a few niche topics such as contour grouping \cite{zhu2007untangling,tversky2004contour,lu2009shape}. It is however interesting and encouraging to notice very recent works on unsupervised 3D reconstruction achieved impressive results by exploring just the rule of symmetry \cite{wu2020unsupervised}.

In this paper, we set out to rejuvenate the research on perceptual organization. Unique to previous attempts that use simple Gestalt rules to explain complicated image data, we (i) approach the problem via purposefully generated \textit{synthetic} data specific to each Gestalt principle, and (ii) ask an agent to solve unseen problems by \textit{generating} the answer from scratch. We encapsulate these design changes to form a novel perceptual organization challenge -- the challenge of perceptual question answering (PQA), where upon observing example PQA pairs, an agent is asked to solve similar perceptual questions by generating answers completely from scratch (see Figure 1 for an example).
  
Our first contribution is therefore a perceptual question and answer (PQA) dataset. Constructing such a dataset is non-trivial as it dictates an explicit measure for perceptual organization -- one needs to design a set of elementary tasks each reflecting a specific Gestalt law. This is in contrast with existing datasets~\cite{linsley2018learning,kim2019disentangling} that are built around real-world scenarios, which do not explicitly model each Gestalt. We are inspired by the recently proposed Abstraction and Reasoning Corpus (ARC) \cite{chollet2019measure}. {In particular, we find the simplicity in forming visual question as color-coded patterns a good match for spelling out specific grouping insights. PQA however comes with two salient features specifically designed for perceptual organization. First, contrary to measuring general intelligence in ARC, PQA gravitates to rule-specific reasoning for perceptual grouping. Second, the size of the hand-crafted ARC dataset can be limited \cite{chollet2019measure}. Instead, our dataset can be deterministically generated thereby producing potentially unlimited number of instances.}

With this dataset, we can train an agent to perform the task of PQA. To realize our agent, we resort to important discoveries in the psychological literature -- that the organization of visual elements could be modeled as an attentive process \cite{mack1992perceptual,yantis1992multielement}. We thus propose a model based on a self-attention mechanism, and formulate the challenge as a grid-to-grid mapping problem. More specifically, our proposed model is built on Transformer \cite{vaswani2017attention} but with three problem-specific extensions: (i) a context embedding module based on a self-attention mechanism to encode an exemplar PQA pair, (ii) positional encoding adapting to the 2D grid nature of our problem, instead of the default 1D case, and (iii) a tailored decoder to predict all symbols (colored blocks on the grid) in parallel, \textit{i.e.}, the entire canvas is produced in one pass to form an answer. Note that unlike conventional visual reasoning tasks which assume a few candidate answers are accessible \cite{santoro2018measuring,zhang2019raven,raven1998raven}, our model directly generates answers from scratch.

Our contribution can be summarized as follows: (i) we rejuvenate the study of perceptual organization through a novel challenge of perceptual question answering, (ii) we propose the first dataset specifically targeting PQA, where each question-answer pair is specific to a particular Gestalt principle, (iii) we formulate perceptual organization via a self-attention mechanism, and propose a grid-to-grid mapping network which is able to directly generate answer-grid from scratch, and (iv) we show our model to outperform a few baselines re-purposed for PQA, \textit{yet} a human study shows that ours uses significantly more data to learn, when compared with an average human. 

\section{Related Work}

\noindent \textbf{Abstract Reasoning} The ability of visual reasoning is fundamental to human-level intelligence \cite{spearman1927abilities,hofstadter1995fluid}, and could be effectively evaluated via IQ tests according to cognitive psychology studies \cite{carpenter1990one,raven1998raven}. Therefore, most works focus on developing relevant datasets and challenges to better comprehend reasoning ability of machines. Typically, candidate answers are accessible for justification, allowing conventional visual reasoning tasks to be treated as a classification problem, such as PGM \cite{santoro2018measuring} and RAVEN \cite{zhang2019raven}. On the contrary, a newly proposed Abstract and Reasoning Corpus (ARC) \cite{chollet2019measure} opens the door for generative reasoning, which requires the subjects to directly {generate answers from scratch}. Similarly, we attempt to motivate a new visual reasoning challenge focusing on perceptual organization, which is particularly formed in a generative manner, \textit{i.e.}, solving a visual question-grid according to a specific Gestalt law, such that it is able to draw an answer-grid from scratch. This would provide a more general benchmark to explore generalization ability of perceptual organization.

\noindent \textbf{Perceptual Question-Answering Dataset} There is a long history of investigating perceptual organization which forms groups of visual stimulus. It is common to use synthetic databases and the corresponding challenges to evaluate perceptual organization/grouping, due to the controllable parameters and outcomes, facilitating an accurate measurement of performance. Pathfinder \cite{linsley2018learning} is a psychology-inspired visual challenge which aims to measure a model's ability of discovering long-range relationship. Different to pathfinder which features low-level cues, cluttered ABC (cABC) challenge \cite{kim2019disentangling} is set to further explore perceptual grouping in a higher object-level. However, none of the existing datasets is designed to measure perceptual organization regarding specific Gestalt laws. We therefore step forward to propose such a new dataset which can facilitate the research in this field.

\noindent \textbf{Perceptual Grouping} The human visual system can naturally perceive the organization of patterns and parts, hence constructing them into meaningful objects. This phenomenon is referred to as perceptual grouping or perceptual organization by psychologists, and the Gestalt theory \cite{king2005max} proposed by Wertheimer Max {stands at the core of} many subsequent vision applications, such as contour grouping \cite{zhu2007untangling,tversky2004contour,lu2009shape,qi2015making}, object detection \cite{gupta2013perceptual,lowe2012perceptual} and image segmentation \cite{gupta2013perceptual,kootstra2011fast}. Nevertheless, major fundamental issues about perceptual grouping still remain unclear, for instance, how to model and measure Gestalt laws precisely, or how to deal with the interaction among multiple principles, \textit{i.e.}, Gestalt confliction \cite{nan2011conjoining,qi2015making,wagemans2012century}. Therefore, in this work, we attempt to facilitate an answer towards a better understanding of the basics about Gestalt laws, by proposing a synthetic dataset (PQA) and a corresponding challenge.

\section{PQA Dataset} 
In this section, we describe the construction of a synthetic visual corpus for perceptual question answering (\emph{PQA}) based on a set of common Gestalt laws. 

\begin{figure}
    \centering
    \includegraphics[width=\linewidth]{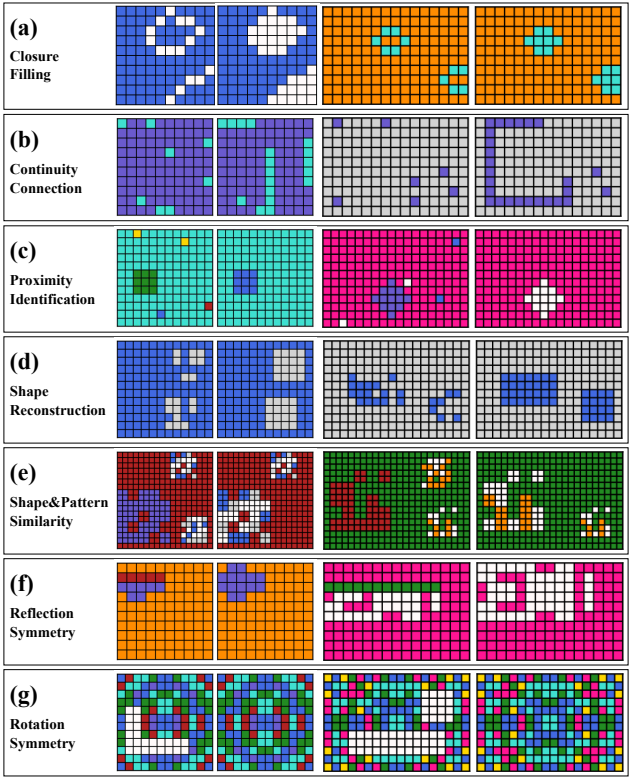}
    \caption{PQA Dataset. Each row from (a) to (g) corresponds to a specific Gestalt law, and a few examples of PQA pair with question (left) and answer (right) are visualized. Zoom in for better visualization. }
    \label{fig:dataset}
\end{figure}

\subsection{Meta Gestalt Laws and Resulting Tasks} \label{tasks}
Seven common Gestalt laws are selected as meta principles to construct our dataset, namely: closure, continuity connection, proximity, shape reconstruction, shape and pattern similarity, reflection symmetry and rotation symmetry. For each law, a corresponding task can be specified as shown in Figure~\ref{fig:dataset} (a)-(g): 
\textbf{T1: Closure Filling} - This task requires one to fill the inner space of a closure shape, with the same color as its surrounding symbol. 
\textbf{T2: Continuity Connection} - Here one needs to discover elements lined up in either vertical or horizontal directions, and fill all positions in between with the same symbol as that of ending-point. 
\textbf{T3: Proximity Identification} - The task needs one to identify the symbol nearest to a shape $S$ that is formed by a group of connected elements, and then replace all elements of $S$ with that symbol (i.e., color $S$ with color of the symbol). Furthermore all other background scattered symbols should be removed as well. 
\textbf{T4: Shape Reconstruction} - Here one is required to reform all irregular shapes into rectangles by filling the least number of intermediate grid slots. The new rectangle should just fit the boundary of the original irregular shapes as shown in Figure~\ref{fig:dataset} (d).
\textbf{T5: Shape Matching \& Pattern Generalization} - Given a triplet ($t, c_1, c_2$) containing a target shape $t$ and two reference shapes $c_1$ and $c_2$, one should determine the reference corresponding to target $t$, and apply its pattern to the target. It should be noted that the candidate and target shapes may be in different scales.
\textbf{T6: Reflection-Symmetry Completion} - Given a set of symbols and a pre-defined axis, one should generate a set of symbols mirroring the existing ones placing the axis as a mirror.  
\textbf{T7: Rotation-Symmetry Completion} - This task is a bit different from others where one needs to fill the \textit{holes} in the question-grid, such that a rotational-symmetric pattern on the answer-grid as whole (Figure~\ref{fig:dataset} (g)).

\subsection{Data format}
\label{sec: dataformat}
Inspired by \cite{chollet2019measure}, we consider each data-tuple $x = (x_q, x_a)$ to have a pair of grid-images -- a question-grid $x_q$ (input) and an answer-grid $x_a$ (output). In raw data representation each question/answer grid of width `$w$' and height `$h$' is composed of $w \times h$ color symbols \footnote{{Note that PQA could be used in a visual format (images) by simply treating Q/A grid as a set of pixels (0-255) rather than color symbols (0-9).}}. A question grid-image can be represented as $x_q = \{s_{i,j}\}_{i=1,j=1}^{w,h}$
where $s_{i,j}$ is a color symbol at location $(i, j)$. Every $s_{i,j}$ is denoted by one of $10$ pre-defined colors in our case. Dimensions ($w,h$) of the grid-image can be of any size up to $30 \times 30$ for our dataset. Availability of such a range of colors and variable grid-space encourages flexibility and variety in designing question-answer grids.

\subsection{Data Synthesis}
\begin{figure}
    \centering
    \includegraphics[width=.95\linewidth]{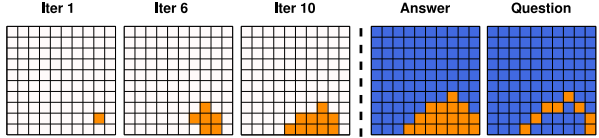}
    \caption{Key steps of data synthesis for T1 (closure filling task).}
    \label{fig:closure}
\end{figure}

The generation of our synthetic dataset \emph{PQA} can be summarized into three steps. First, we consider a blank grid canvas with random size $w\times h \leq 30 \times 30$. Secondly, patterns of answer-grid are generated by obeying a specific Gestalt law, and thirdly, question-grid is obtained by altering its corresponding answer-grid -- as in deleting or replacing symbols (colored blocks on the grid). Figure \ref{fig:closure} shows the grid-image generation process for T1 (closure filling): (i) Generate a blank canvas for grid-image. (ii) {Starting from a random location, expand vertically or horizontally by repetitively placing the same symbol in any one of the 8 slots linked to the current location. After $k$ iterations, a foreground formed from connected symbols is obtained as an answer-grid.} (iii) To obtain the question-grid, we remove the internal symbols of the closure region leaving behind only the boundary symbols. A total of 100k PQA pairs per task constitute our dataset. {Note that the answer for each question is unique, which is guaranteed by the data generation procedure. Further details of data generation and data formats are available in supplementary material.}

\begin{figure}
    \centering
    \includegraphics[width=.95\linewidth]{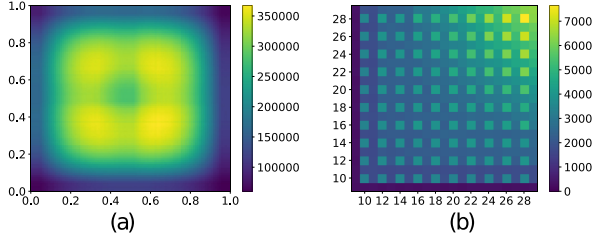}
    \caption{Data distribution. (a) Distribution of key region locations. x and y coordinates are normalized to (0,1), corresponding to the center of key regions. (b) Distribution of grid size. x-axis and y-axis correspond to width and height of a grid.}
    \label{fig:statistic}
    \vspace{-0.5cm}
\end{figure}

\subsection{Context PQA pair and Test Question}
{During inference of the model, two components are provided - a context PQA pair and a test question. With the help of the given context PQA pair, the model infers the corresponding Gestalt law used there. Subsequently this inferred law is used to generate a new answer-grid corresponding to the test question from scratch.This design tests two essential abilities of the model, (i) abstract reasoning skill, using which the model extracts the implicit Gestalt principle in the visual sample, and (ii) its generalization capacity, using which it applies the learned rule to a new question and \emph{generates} the relevant answer directly. }

\subsection{Statistical Analysis}
To gain further insights into our dataset, statistical analysis is provided as shown in Table \ref{tab:dataset} and Figure~\ref{fig:statistic} where \textit{Avg Symbols} indicate the number of symbols in a question-grid. It basically shows the number of colors that are enough to represent a specified instance of a task. For instance, 2 colors in T1 are enough -- one for background and one for the boundary of closure region.
\textit{Avg Slots} represents the percentage of question-grid needed to be \textit{modified} to form a correct answer.
We observe the following from Table \ref{tab:dataset} : (i) Due to the varying pattern complexity, the number of symbols within an image-grid is different. (ii) Tasks T1, T2, T4 and T6 typically use fewer color symbols, while others require more. (iii) \emph{Avg Slots} shows that task T2 (continuity connection) and task T3 (proximity identification) require less editing while task T5 needs the most. Judging from this perspective, task T5 seems more complicated than the others.
{We also observe from Figure~\ref{fig:statistic}(a) that key regions (foreground) are evenly distributed on the canvas and are usually located near the center. Furthermore, it should be noted that due to difference in size (Figure~\ref{fig:statistic}(b)) of different question-answer grid pairs, the difficulty level for generating answers from scratch is variable. Importantly, the greater the size of the answer-grid to be generated, the higher is the possibility of error, which in turn reflects the flexible and challenging nature of the dataset.  }

\begin{table}[t] 
    \caption{Statistics of PQA dataset.}
    \centering
    \small
    \begin{tabular}{lccccccc}
    \hline
    \hline
       Tasks  & $T_1$ & $T_2$ & $T_3$ & $T_4$ & $T_5$ & $T_6$ & $T_7$  \\
       \hline
       Avg Symbols & 2.0 & 2.0 & 5.0 & 2.0 & 5.0 & 3.0 & 5.0  \\
       Avg Slots (\%) & 12.9 & 3.6 & 4.0 & 7.6 & 15.3 & 9.8 & 12.5
       \\
       \hline
       \hline
    \end{tabular}
    \label{tab:dataset}
    \vspace{-0.6cm}
\end{table}

\section{Methodology}
\begin{figure*}
    \centering
    \includegraphics[width=\linewidth]{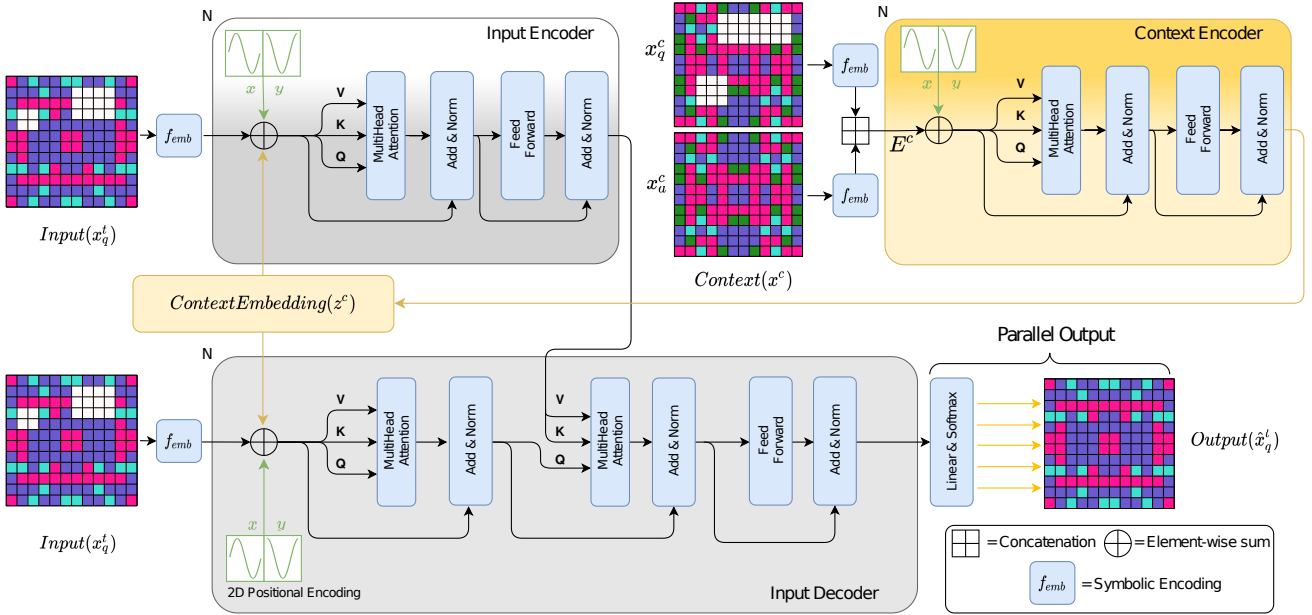}
    \caption{Our framework. The encoder takes test question embedding, positional encoding and context embedding as inputs, where context embedding is given by a context encoder, providing clues about the implied law in an example PQA pair, and the positional encoding adapts to the 2D case. The decoder can generate an answer-grid by predicting all symbols at every location in parallel. }
    \label{fig:network}
\vspace{-0.3cm}
\end{figure*}

We propose a Transformer-based generative model to learn perceptual abstract reasoning. The goal is to directly generate an answer image-grid according to a visual question-grid of an unknown task, with the help of an exemplar question-answer pair that is provided as context for inferring the implicit Gestalt law.
Formally, given a context grid pair $x^c = (x^c_q, x^c_a)$ and the test question grid $x^t_q$, our model is to generate an answer grid $x^t_a$ by estimating a color symbol at each grid-location from scratch.

\subsection{Model Architecture}
As shown in Figure~\ref{fig:network}, our proposed network is an encoder-decoder architecture which is tailored from Transformer~\cite{vaswani2017attention}. In general, the encoder is a stack of N identical layers, and each layer takes inputs from three sources: (i) test question embedding (the first layer) or output of last layer (the other layers), (ii) positional encoding and (iii) context embedding.
The decoder then generates an answer-grid by predicting all symbols at every location of the grid. There are three fundamental components in our transformer architecture: (i) A context embedding module based on self-attention mechanism that is used to encode exemplar PA grid pairs which itself is a part of the input fed into the input encoder. This plays a significant role in inferring the implicit Gestalt law and later generalizing onto the new test question. (ii) We extend positional encoding to adapt to 2D grids-case instead of working on 1D sequence-case, as the 2D locations of symbols are of crucial importance for our problem. (iii) As ours is a 2D grid-to-grid mapping problem, the decoder is tailored to predict all symbols in parallel, i.e., all the color grids are produced in one pass to form an answer instead of one output at a time. Further details are provided in the following sections.

\subsubsection{Encoder and Decoder}

\keypoint{Encoder: } 
{Following transformer architecture in \cite{vaswani2017attention}, the encoder is designed as a stack of N identical layers, each of which has two sub-layers, i.e., a sub-layer of multi-head self-attention (MHA) and a feed-forward network (FFN). We define a single-head self-attention (SHA) module $\texttt{attn}(Q,K,V)$ as:}
\begin{equation}
    \texttt{attn}(Q,K,V) = \texttt{softmax}(\frac{QK^T}{\sqrt{d_k}})V
\end{equation}
Accordingly the other components of the network are defined as:
\begin{equation}
\label{mha}
\begin{aligned}
    \mathrm{SHA}_i(Q,K,V) &= \texttt{attn}(QW^Q_i,KW^K_i,VW^V_i)  \\
    \mathrm{MHA}(Q,K,V) &= \texttt{concat}(\mathrm{SHA}_1, ...,\mathrm{SHA}_h)W^O 
\end{aligned}
\end{equation}
\begin{equation}
    \mathrm{FFN}(x) = \max(0, xW_1 + b_1)W_2 + b_2
\end{equation}
{where Q, K and V are \emph{Query}, \emph{Key} and \emph{Value} respectively, $\mathrm{SHA}_i$ is the $i$-th attention head. FFN ($\cdot$) is composed of two linear transformations with ReLU in between where {$W_{(\cdot)}^{(\cdot)}$ are learnable matrices.} Around each of the sub-layers, a residual connection with layer normalization is placed. Therefore, the output of each sub-layer is $\texttt{LayerNorm}(x+\texttt{sublayer}(x))$.}

\keypoint{Decoder: } The decoder is also designed as a stack of N identical layers following \cite{vaswani2017attention} with 3 sub-layers, -- 2 MHA sub-layers followed by 1 FFN. The first MHA sub-layer learns to attend to the test question itself, while the inserted additional MHA sub-layer performs \textit{self-attention} on the output of the encoder. Intuitively, the former MHA reads the question, while the latter MHA reads the context. Similar to encoder stack, residual connections and layer normalizations are applied to each sub-layer as well. Notably, our decoder can attend to the whole question grid, and can therefore generate answer grid by estimating symbols at each location in parallel, with the help of linear transformations and softmax.

\subsubsection{Input Representation}
{The representation of any input grid to encoder or decoder is obtained by element-wise summation of its symbolic embedding and positional embedding tensors}.

\keypoint{Symbolic embedding: } 
Given a grid in original format {$x \in \mathbb{R}^{w\times h}$}, each entry $x_{i,j}$ is assigned with a symbol ranged from 0 to 9, where each number corresponds to an unique color in our case as mentioned before (\S~\ref{sec: dataformat}). Following \cite{vaswani2017attention}, we use a learned embedding function $f_{emb}$ to map each symbol $x_{i,j}$ to a vector of dimension $d$. Consequently, an embedding $f_{emb}(x)$ is obtained for $x$ as: 
\begin{equation}
    E_x = f_{emb}(x) \; ; \quad E_x \in \mathbb{R}^{w\times h\times d}
    \label{equ: symbolic}
\end{equation}

\keypoint{Positional embedding: } As positional information of symbols is naturally crucial for reasoning on an image-grid, we choose to encode such information into our network. Unlike sequential data (\textit{e.g.} sentences), our input is an image-grid where both coordinates at $(i,j)$ need to be encoded. Accordingly we encode co-ordinates $i$ and $j$ as $\mathrm{PE}_i$ and $\mathrm{PE}_j$ respectively and concatenate them to obtain our positional embedding as: $\mathrm{PE}_{(i,j)} = \texttt{concat}(\mathrm{PE}_i, \mathrm{PE}_j) \in \mathbb{R}^{d}$. Formally, position $pos$ can be encoded using $sine$ and $cosine$ functions with different frequencies as:
\begin{equation}
\begin{aligned}
    \mathrm{PE}_{pos}^{2k} &= \sin{(pos/10000^{2k/d})} \\
    \mathrm{PE}_{pos}^{2k+1} &= \cos{(pos/10000^{2k/d})}
\end{aligned}
\end{equation}
where $k$ is the dimension and $pos$ could be $i$ or $j$. In contrast to Transformer which only takes positional encoding at the first encoder/decoder layer, we demonstrate via experiments that injecting positional embedding to every encoder/decoder layer is beneficial towards performance.

\subsubsection{Context Embedding: }
Apart from the test question $x$ in input, context $x^c$ is the only information source of understanding the implicit Gestalt law. The exemplar PQA pair $x^c = (x^c_q, x^c_a)$ thus also needs to be encoded. 
Accordingly we first obtain the symbolic embedding of question ($x^c_q$) and answer ($x^c_a$) grids as $E^c_q$ and $E^c_a$ respectively following Eq. \ref{equ: symbolic}. They are then concatenated together as $E^{c} = \texttt{concat}(E^c_q, E^c_a)$ which is then fed into a \textit{context encoder} (same architecture as that of input encoder) to produce a latent representation of $z^c \in \mathbb{R}^{d}$ as the final context embedding.

\vspace{-0.2cm}
\subsubsection{Objectives: }
Given the decoded output of $\hat{x}^t_a \in \mathbb{R}^{w\times h}$ where each entry $\hat{x}^t_a(i,j)$ denotes the predicted symbol at output grid location $(i,j)$, we treat this as a multi-class classification problem. A negative log likelihood loss ($\mathcal{L}_{cls}$) is used to measure whether a correct color symbol is estimated at each location on the grid as:
\vspace{-0.3cm}
\begin{equation}
    \mathcal{L}_{cls} = - \sum_i^w \sum_j^h x^t_a(i,j) \; \operatorname{log}(\hat{x}^t_a(i,j))
\vspace{-0.3cm}
\end{equation}
where $x^t_a$ is the ground-truth answer-grid.

\section{Experiments}
\subsection{Experimental settings}
\keypoint{Datasets:} {We use our introduced \textit{PQA} dataset for all experiments.} For each of the seven tasks, there are 100k PQA grid-pairs, which are split into training and testing set in the ratio of 4:1, where half of the PQA pairs in either training/testing split will be used as context PQA pairs. 

\keypoint{Implementation details:} 
Our implementation follows Transformer \cite{vaswani2017attention} in practice. Both the encoder and decoder have 6 identical layers stacked, and 8 parallel heads are employed in each layer. The output dimension of each layer is $(b,w,h,512)$, where the batch size $b$ is set to 16 with dimension of each symbol set to $512$. We have 10 color symbols available for constructing a large Gestalt law-specific database. We additionally use a ``padding'' symbol to pad all data within a mini-batch to the same size for implementation purpose. Adam optimizer with $\beta_1 = 0.9$, $\beta_2 = 0.99$ and $\epsilon=10^{-8}$ is employed. Learning rate is set to $0.0001$ with decay rate $0.1$ for every epoch, which contains 3500 randomly chosen batches. The same residual dropout strategy with a rate of $0.1$ is used. Our model is implemented by PyTorch \cite{paszke2017automatic} trained using a single Nvidia Tesla V100 GPU card.

\keypoint{Evaluation protocol:} We evaluate by measuring the percentage of error-free answers, where only answers with \textit{all} the symbols correctly estimated, are counted.

\keypoint{Competitors:} On absence of existing methods for perceptual abstract reasoning, we compare our method against several alternative state-of-the-art networks, by adjusting to generative models, including \textbf{ResNet} \cite{he2016deep}, \textbf{LSTM} \cite{hochreiter1997long}, \textbf{Transformer} \cite{vaswani2017attention}, and \textbf{TD+H-CNN} \cite{kim2019disentangling}. 
{It is worth mentioning that we train \textit{all} the models \textit{from scratch} on our synthetic PQA dataset.}
\textbf{ResNet} is explored to testify the performance of CNN-based model on perceptual abstract reasoning, our proposed variant of ResNet is composed of a symbolic embedding layer to represent data in original format, a stack of 16 ``BasicBlock'' used in \textbf{ResNet-34}, followed by up-sampling and softmax to output symbol probabilities at each location to form the answer-grid. Similarly, we also adopt deeper architecture of \textbf{ResNet-101} with a stack of 33 ``bottleneck'' building blocks replacing all the ``BasicBlock'' as backbone for evaluation. Notably, the PQA context is also integrated into network by concatenating its feature embedding with its counterpart of input question along channel. 
\textbf{LSTM} stands for a representative of RNN-based model. Here 2D grids can be directly converted to 1D sequences via `flatten' transformation, thus testing potential of RNN in our problem. In practice, the same symbolic embedding layer is applied to input and context grids first. They are then concatenated with the channel dimension, followed by a flatten transformation to form the sequence of feature vectors. \textit{i.e.} each point corresponds to a symbol, which is fed into a typical LSTM. The final hidden state vector is transformed by two fc-layers with softmax operation for symbol prediction. We further explore the performance of bi-directional LSTM (\textbf{bi-LSTM}) by feeding the input sequence from both directions.
\textbf{Transformer} shares the same core network architecture as ours. To gauge the original Transformer model's efficiency we simply enable the decoder to attend all grid locations instead of using masked multi-head attention mechanism. To facilitate context encoding, the same context embedding module of our network is used. The obtained context embedding is then added to input embedding and the positional encoding, to form the final input to the encoder. \textbf{TD+H-CNN}, which is a state-of-the-art method on perceptual grouping, implements bottom-up, horizontal and top-down connections in convolutional neural networks. It was originally used for grouping any two points in a given image, by producing a binary category estimation.  We tailor it for dealing with our task by modifying the readout module into a multi-category classifier to predict the color symbol at each location of the answer-grid.

\begin{table}

\setlength{\tabcolsep}{2.5pt}
    \centering
    \normalsize
    \caption{Comparison results ($\%$) of models trained on \emph{all tasks}.}
            \begin{tabular}{l|cccccccc}
                \hline
                \hline
                Method  & $T_1$ & $T_2$ & $T_3$ & $T_4$ & $T_5$ & $T_6$ & $T_7$ & Avg  \\
                \hline
                ResNet-34  & 79.6 & 17.6 & 17.9 & 85.2 & 0 & 19.6 & 0.1 & 31.4 \\
                ResNet-101  & 73.9 & 10.6 & 0.1 & 50.9 & 0 & 1.7 & 0 & 19.6 \\
                LSTM  & 55.7 & 23.2 & 25.6 & 38.2 & 0 & 7.4 & 2.8 & 21.8\\
                bi-LSTM  & 81.9 & 26.6 & 75.6 & 85.9 & 0 & 41.4 & 23.4 & 47.8\\
                Transformer  & 16.8 & 11.3 & 87.4 & 0.3 & 0 & 0.1 & 0 & 16.7 \\
                TD+H-CNN & \bf 88.8 & 89.8 & 78.8 & 96.4 & 0 & 50.8 & 9.3 & 59.1\\
                Ours & 82.6 & \bf 97.6 & \bf 93.7 & \bf 96.9 & \bf61.8 & \bf82.7 & \bf98.9 & \bf 87.8 \\
                \hline
                \hline
            \end{tabular}
    \label{tab:overall}
    \vspace{-0.2cm}
\end{table}

\begin{figure}
    \centering
    \includegraphics[width=.85\linewidth]{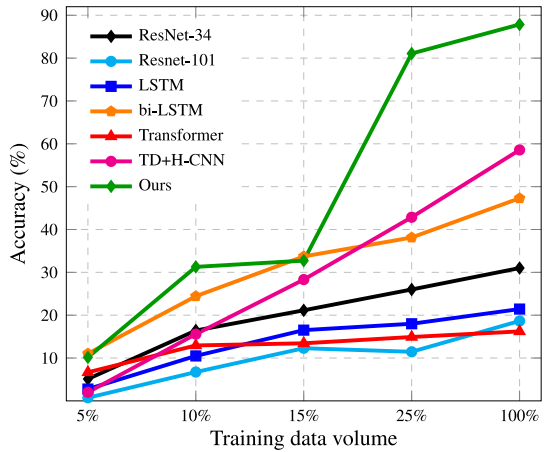}
    \caption{Testing results on varying training data volume. }
    \label{fig:datascale}
    \vspace{-0.2cm}
\end{figure}

\subsection{Results}
\begin{figure}
    \centering
    \includegraphics[width=.95\linewidth]{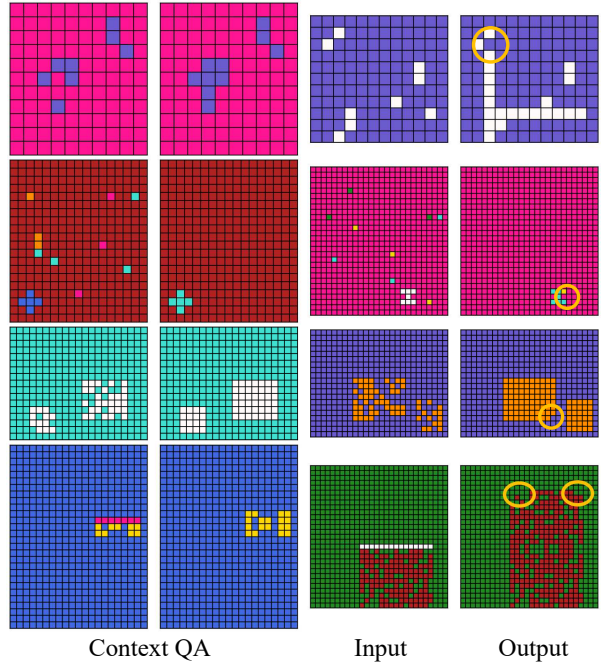}
    \caption{Failure cases. Errors are marked in yellow circle.}
    \label{fig:results}
    \vspace{-0.5cm}
\end{figure}

Table~\ref{tab:overall} shows quantitative results of every method trained on all the tasks in our \emph{PQA} dataset. We observe our method to secure an average precision of $87.8\%$, which significantly outperforms other competitors over all tasks. On inspecting performances of every task further individually in Table~\ref{tab:overall}, we realize that T5 is most challenging as all baseline methods fail completely. On the contrary, it is interesting to note that humans can understand and address the questions in T5 quite easily, {as shown later in Table~\ref{tab:human}}. Similar trend can be found on T6 and T7 as well. On tasks T1 to T4, all competitors perform better than they do on T5 to T7. Furthermore TD+H-CNN achieves result comparable to ours on T4. To further evaluate the training efficiency of each model, we provide different amounts of data for training. We can observe from Figure~\ref{fig:datascale} that the scale of training data significantly affects model's performance. Unlike our model, humans can learn the task-specific rule from very limited examples {(Table~\ref{tab:human})}. Basically all methods  would nearly fail (lower than $15\%$) if we reduce the amount of training data to $5\%$ of PQA pairs per task. Compared to other baseline methods however, ours performs the best. We additionally showcase some failure cases produced by our model in Figure~\ref{fig:results}, a typical case is that only a tiny fraction of symbols (maybe only one) are not correct, resulting in a completely false answer.

\subsection{Ablation study} 

To justify the efficiency of every key component : context embedding and positional embedding, we conduct an in-depth ablation study where each component is stripped down at a time to inspect relative contribution (Table~\ref{tab:abla}).

\keypoint{Is \textit{context} important?:} To answer this we remove context embedding module (\textbf{w/o Context}) from our model, which means the context PQA pairs are now unavailable. {Interestingly, our model can still achieve $65.8\%$ without using context, this may be due to useful knowledge already being established that can discriminate different tasks after observing \emph{enough} data during training. However, a drastic fall in average accuracy from $87.8\%$ to $65.8\%$ also suggests that context embedding is indeed beneficial to reasoning the implied law and in distinguishing among different tasks. }

\keypoint{Essence of positional encoding in every attention head:} To judge the utility of injecting positional encoding to every multi-head attention (MHA) module, we switch back to the original setting which embeds positional encoding to the \textit{first} MHA module \textit{only} (\textbf{Front-{PE}} \& \textbf{w/o context}). A performance collapse of $47.7\%$ justifies our design choice.

\begin{table}
\small
    \caption{Ablation study (\%).}
    \setlength{\tabcolsep}{3pt}
    \centering
    \begin{tabular}{ccccccccc}
    \hline
    \hline
      Variants  & $T_1$ & $T_2$ & $T_3$ & $T_4$ & $T_5$ & $T_6$ & $T_7$ & Avg \\
      \hline
        Front-\text{PE} \&   & \multirow{2}{*}{26.1} & \multirow{2}{*}{10.5} & \multirow{2}{*}{89.4} & \multirow{2}{*}{0.4} & \multirow{2}{*}{0} & \multirow{2}{*}{0.2} & \multirow{2}{*}{0} & \multirow{2}{*}{18.1}\\
        w/o Context & & & & & & & &\\
        w/o Context  & 69.5 & 94.9 & 87.6 & 91.7 & 0 & 37.2 &79.8 & 65.8 \\
        Ours & \bf82.6 & \bf97.6 & \bf93.7 & \bf96.9 & \bf61.8 & \bf82.7 & \bf98.9 & \bf87.8 \\
        \hline
        \hline
    \end{tabular}
    \label{tab:abla}
    \vspace{-0.4cm}

\end{table}

\begin{figure}[t!]
    \centering
    \includegraphics[width=\linewidth]{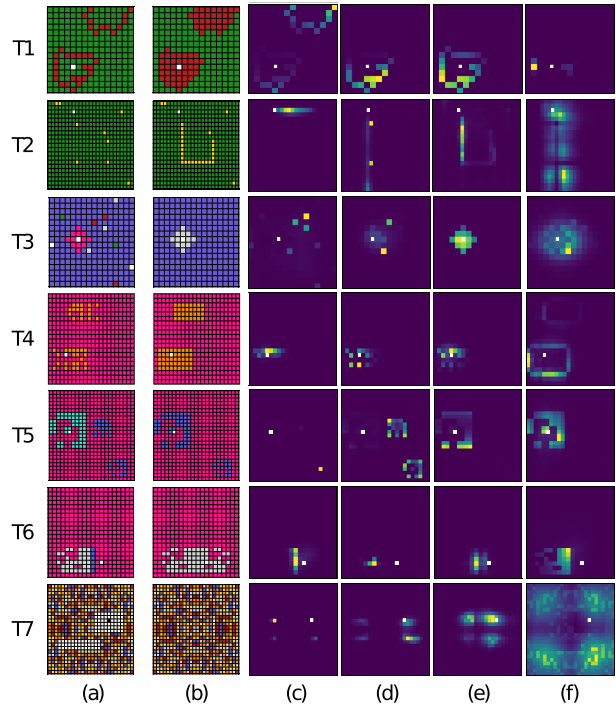}
    \caption{Exemplar attention maps from second MHA sub-layer of the decoder. (a) Question. (b) Generated answer. (c)-(f) Attention maps from different heads for a pre-determined grid location, which is denoted by a white dot or black dot for clarity.}
    \label{fig:attention}
    \vspace{-0.2cm}
\end{figure}

\noindent \textbf{Why Attention?: } 
To gain further insights on \textit{where} the network would attend to while generating a symbol at a grid location, we further visualize the attention maps of the decoder's second MHA sub-layer, in Figure~\ref{fig:attention}. Interestingly our model can not only spot foreground grids but also searches nearby background regions -- a human-like behavior. As we can see from T1's attention map (d) and (e) it can simultaneously attend to the closure boundary as well as its nearby outside regions when generating a symbol at the location denoted by white dot. 
Taking T3 as example, the model can locate target regions in (e) and (f), and compare the scattered dots in (c) and (d), thus demonstrating its reasoning ability. Similar observation is evident from other tasks as well.

\begin{table}[h]
\setlength{\tabcolsep}{3pt}
    \caption{Average number of contextual PQA pairs needed to fully understand the tasks by humans.}
    \centering
    \begin{tabular}{lccccccc}
        \hline
        \hline
        Tasks & $T_1$ & $T_2$ & $T_3$ & $T_4$ & $T_5$ & $T_6$ & $T_7$  \\
       \hline
       \# context & 1.5 & 1.75 & 3 & 1 & 1 & 1.25 & 1  \\
       Avg time & 2.48 & 2.67 & 5.93 & 0.77 & 3.95 & 2.54 & 1.65  \\
       \hline
       \hline
    \end{tabular}
    \label{tab:human}
\vspace{-0.4cm}
\end{table}
\section{Human Study}
\vspace{-0.1cm}

{Humans are highly adept at visual reasoning \cite{hofstadter1995fluid}, and are equipped with a remarkable perceptual ability in understanding abstract and incomplete structural cues. Many IQ-test driven abstract reasoning challenges, such as PGM \cite{santoro2018measuring}, RAVEN \cite{zhang2019raven} and ARC \cite{chollet2019measure}, are used to benchmark AI intelligence. It has been shown that humans could learn from a very limited number of exemplar QA pairs to address new tests. To gauge how closely our model understands perceptual reasoning, we need some insights on how humans perceive and accomplish similar tasks on our PQA dataset. }

{Accordingly, a human study is conducted using the same user interface as ARC \cite{chollet2019measure}. A total of $10$ people were recruited. For every Gestalt law, each participant is required to browse as \textit{few} exemplar PQA pairs as possible, until they are satisfied that they are ready to infer new visual questions on that law.  We then ask them to answer 3 \textit{consecutive} questions correctly to verify that they had \textit{truly} learned the law, and discard those cases with a failed test. We record the number of contextual pairs each examined, and the time (in mins) spent on each task. As expected (Table~\ref{tab:human}), humans need just a few PQA instances to learn the implied rule from shown examples. In fact humans can accomplish those same task with astronomically fewer data ($<$2 samples) than our agent, and \textit{always} generate correct answers. This clearly signifies just how unexplored this topic is, and in turn encourages future research to progress towards human-level intelligence.}


\section{Conclusion}
{In this paper, we set out to rejuvenate research on perceptual organization. We for the first time frame it as a perceptual question-answering (PQA) challenge, that requires an agent to infer an implied Gestalt law with the help of an exemplar PQA pair and then generalize to a new test question. Importantly, we propose a synthetic dataset for perceptual question-answering (PQA) that consists of Gestalt-specific PQA pairs. We then introduce the first PQA agent that models perceptual organization via a self-attention mechanism, and show that it can reliably solve PQA with an average accuracy of $87.76\%$. A further human study however revealed that our model needs astronomically more training data to learn when compared with humans, signifying the need for future research. We will release the dataset and make our code publicly accessible. Last but not least, we hope this dataset will trigger a renewed wave of research on perceptual organization, and help to spell out relevant insights that could be of benefit to other vision tasks \cite{wu2020unsupervised}.}

\section*{Acknowledgement}
This work was supported by the National Natural Science Foundation of China (NSFC) under 61601042.

{\small
\bibliographystyle{ieee_fullname}
\bibliography{egbib}
}

\end{document}